\documentclass{article}

\usepackage{microtype}
\usepackage{graphicx}
\usepackage{subcaption}
\usepackage{booktabs} 
\usepackage[utf8]{inputenc}
\usepackage{titling}

\usepackage{hyperref}

\usepackage[accepted]{icml2026}

\usepackage{amsmath}
\usepackage{amssymb}
\usepackage{mathtools}
\usepackage{amsthm}

\usepackage[most]{tcolorbox}
\usepackage{listings}
\usepackage{xcolor}

\definecolor{lightgrey}{gray}{0.95}
\definecolor{darkgrey}{gray}{0.4}

\usepackage[capitalize,noabbrev]{cleveref}

\theoremstyle{plain}

\theoremstyle{definition}

\theoremstyle{remark}

\usepackage{algorithm}
\usepackage{algorithmic}                                                                                                                                                                    
\usepackage[textsize=tiny]{todonotes}

\icmltitlerunning{\raf: Scaling Long Horizon Agentic Fault Attribution via Active Investigation}

\usepackage{xspace} 

\newcommand{\raf}{\textsc{SAFARI}\xspace}

\begin{document}

\twocolumn[
\icmltitle{
\raisebox{-0.25\height}{\includegraphics[height=1.1cm]{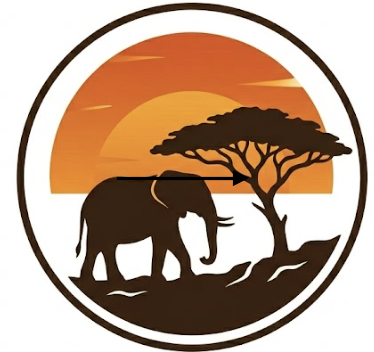}}
\hspace{0.25cm}
SAFARI: Scaling Long Horizon Agentic Fault Attribution \\
via Active Investigation
}



  \icmlsetsymbol{equal}{*}
\begin{icmlauthorlist}
    \icmlauthor{Chenyang Zhu}{c1}
    \icmlauthor{Jiayu Yao}{c1}
    \icmlauthor{Kushal Chawla}{c1}
    \icmlauthor{Youbing Yin}{c1}
    \icmlauthor{Nathan Wolfe}{c1}
    \icmlauthor{Pengshan Cai}{c1}
    \icmlauthor{Jingyu Wu}{c1}
    \icmlauthor{Spencer Hong}{gic,nu}
    \icmlauthor{Sangwoo Cho}{c1}
    \icmlauthor{Shi-Xiong Zhang}{c1}
    \icmlauthor{Daben Liu}{c1}
    \icmlauthor{Sambit Sahu}{c1}
    \icmlauthor{Erin Babinsky}{c1}
\end{icmlauthorlist}

\icmlaffiliation{c1}{AI Foundations, Capital One}
\icmlaffiliation{gic}{General Intelligence Company of New York}
\icmlaffiliation{nu}{Department of Engineering Sciences and Applied Mathematics, Northwestern University}

\icmlcorrespondingauthor{Chenyang Zhu}{chenyang.zhu@capitalone.com}

  \icmlkeywords{Machine Learning, ICML}

  \vskip 0.3in
]



\printAffiliationsAndNotice{}  

\begin{abstract}
As autonomous agents tackle increasingly complex multi-step, multi-agent tasks, their execution trajectories have scaled beyond the constraints of even the largest context windows. 
Current methods for effectively diagnosing agent failures load the full trajectory into an LLM's context window, which suffers from attention dilution and fails when agentic traces inevitably exceed context limits. To address this, we introduce \raf (\textbf{S}caling long-horizon \textbf{A}gentic \textbf{F}ault \textbf{A}tt\textbf{R}ibution via active \textbf{I}nvestigation), a framework that replaces linear context loading with a tool-augmented diagnostic loop. By equipping LLMs with a specialized toolbox to \texttt{read} and \texttt{search} trajectory segments alongside a persistent Short-Term Memory (STM) for cross-turn reasoning, SAFARI effectively decouples diagnostic accuracy from architectural context limits. Our experiments demonstrate that \raf outperforms state-of-the-art results by 20\% on the Who\&When dataset within a 1M token budget, and by 19\% on TRAIL GAIA subset on a 25K token budget. Most significantly, \raf maintains a 0.58 precision even when the target fault resides 5x beyond the model’s native context window, a scenario where traditional evaluators fail entirely. 

\end{abstract}

\section{Introduction}
The rapid advancement of Large Language Model (LLM)-driven agentic systems has enabled the automation of increasingly complex tasks. As these systems scale toward multi-agent coordination, it becomes correspondingly important to diagnose where and how these systems fail \cite{Cemrietal2025}. Fault attribution, the task of isolating the specific agent or step responsible for a failed trajectory, is essential for system monitoring and reliability. 

Within the field of fault attribution, it has become evident that human annotation and single-LLM judges are unable to accurately and efficiently detect faults in multi-agent systems \cite{whowhen}. Over the past year, two classes of fault attribution capabilities have been introduced: (a) simulation-based methods that run partial or entire trajectories to locate faults \cite{agentdebug, epperson2025interactive, Geetal2025}; and (b) simulation-free methods that leverage multi-agent invocations, iterative loops, and tool calls \cite{zhu2026raffles, yu2025correct}.

However, challenges remain for state-of-the-art fault detectors to correctly identify faults over long horizons. To date, fault detection systems show notable drops in performance on the Who\&When \cite{whowhen} and TRAIL \cite{deshpande2025trail} benchmarks when trajectory lengths increase \cite{zhu2026raffles, deshpande2025trail}. In TRAIL particularly, the token lengths of agentic trajectories have reached multiple millions of tokens, well exceeding the context windows of open-source and proprietary large language models. Given the limitations of current systems, it is important to introduce fault detector solutions that can scale, cost-effectively, to these new requirements. 

Current strategies for handling long horizons include utilizing agentic reflection-based reasoning \cite{zhu2026raffles} and massive 1M-token context windows \cite{deshpande2025trail, agentdebug}. However, these approaches have the limitation that they process entire trajectories passively for fault detection. Recent literature reveals that, instead, active investigation (i.e. strategically storing and retrieving key information) can outperform ingesting the entire context as input for for long-horizon tasks. 
Frameworks like AgentFold \cite{ye2025agentfold} and LIGHT \cite{tavakoli2025beyond} demonstrate this by using explicit read-write skills to ensure critical details remain actionable within massive information stores.

We propose \raf, a fully autonomous Active Investigator designed to localize errors within agentic traces through iterative, hypothesis-driven exploration. Unlike traditional models that passively ingest entire trajectories, \raf dynamically interacts with the trace by generating atomic hypotheses, deploying specialized search tools, and synthesizing expert judges to verify its own findings. We evaluate \raf across two major benchmarks: Who\&When \cite{whowhen} and TRAIL \cite{deshpande2025trail}, providing a holistic assessment of fault attribution across varying horizons. Our results demonstrate that \raf significantly outperforms existing methods in resource-constrained environments, effectively identifying faults that reside beyond the native context limits of current large language models.

\section{Methodology: \raf}

\begin{figure*}[t] 
    \centering
    \includegraphics[width=.95\linewidth]{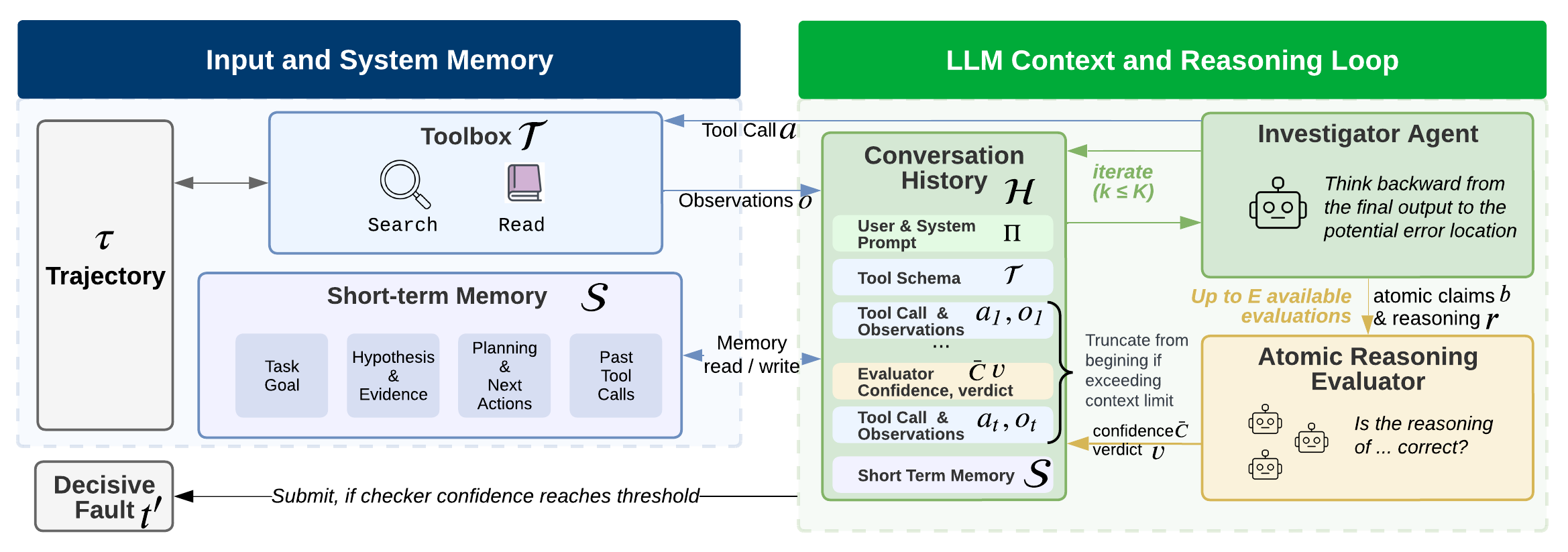}
    \caption{Illustration of \raf's Active Investigation loop. }
    \label{fig:illustration}
\end{figure*}

Given a trajectory $\tau=\{\tau_1, \tau_2, \dots, \tau_T\}$ from an agentic system, where each step $\tau_t$ comprises an agent’s input and output, the fault attribution task seeks to identify a decisive fault $t^* \in [1, T]$, defined as the earliest uncorrected error within the trace. Rather than ingesting the entire trajectory into the LLM’s limited context window, SAFARI employs an Investigator Agent. This agent interacts with $\tau$ through iterative tool calls, selectively querying trajectory segments to gather evidence before reaching a diagnostic conclusion.

Each investigation begins by initializing the conversation history $\mathcal{H}_0=\{\Pi_{\text{sys}}, \Pi_{\text{usr}}\}$. The system prompt, $\Pi_{\text{sys}}$, defines the fault attribution objectives and instructions. This is followed by the user prompt, $\Pi_{\text{usr}}$, which grounds the agent by providing the total trace length, the initial investigation metadata, and the data structures of the agentic system being evaluated. 


At each iteration $k$, the Investigator Agent observes the current history $\mathcal{H}_k$, which incorporates all prior tool calls and their observations, and the tool schema $\mathcal{T}$ to produce tool calls $a$. Concretely, $a$ is drawn from two specialized tool categories designed to navigate the vast search space of trajectory $\tau$:

$\texttt{read}(\textit{offset}, \textit{limit})$ systematically traverses trace segments, mimicking human debugging where an engineer ``scrolls'' through logs surrounding a suspected failure point; $\texttt{search}(\textit{pattern})$ executes case-insensitive regex queries over the serialized trajectory, allowing the agent to instantly ``jump'' to specific span IDs, variable values, or known error signatures without exhausting the context window.

Once sufficient evidence has been gathered, \raf\ enforces a verification stage before the Investigator Agent can conclude. The Investigator Agent decomposes its hypothesis into up to $E=3$ atomic claims $b_i$ with reasoning $r$. Each claim serves as a falsifiable statement about a specific step (e.g., ``span \texttt{abc123} stated value $X$ without any tool call in its input''). The claims are dispatched in parallel to an Reasoning Evaluator LLM, which verifies \emph{only that one claim} against the quoted evidence in the reasoning. The evaluators have no access to the underlying trace; they evaluate solely whether the cited evidence supports the stated conclusion and outputs a confidence score for the evaluation. The average confidence score $\bar C \in [0, 100]$ and per-claim feedbacks $\{v\}$ are returned to the Investigator Agent. If $\bar C \geq \theta$ (default $\theta = 70$), the Investigator Agent is prompted to conclude and output the decisive fault. Otherwise, the low-scoring claims pinpoint exactly which logical steps require stronger evidence, and the tool call $a$ and observation $o$ are appended to the history to form $\mathcal{H}_{k+1} = \mathcal{H}_k \cup \{(a, o)\}$.



The accumulated history $\mathcal{H}_k$ forms a \textit{chain-of-evidence} that conditions subsequent queries. However, as $\mathcal{H}_k$ grows, the oldest turns are silently evicted once the context budget is exhausted, erasing evidence that was expensive to gather. To address this, \raf\ maintains a \textbf{short-term memory} (STM) $\mathcal{S}$ that is appended \emph{at the end of} $\mathcal{H}_k$ on every LLM call. The STM has four main fields: underlying task goal, hypothesis and evidence of failure steps, investigative gaps that store plans and signal next actions, and past tool calls. 

The investigation loop terminates upon a valid conclusion with the final prediction, $t'$, and the supporting rationale, $R'$, or when the iteration budget $K=30$ is exhausted. In cases of exhaustion, \raf\ executes a ``forced finalization'' routine, where the Investigator Agent is prompted to synthesize its best answer based on the evidence gathered in the history $\mathcal{H}_K$. The complete \raf framework is depicted in \Cref{fig:illustration}. We exhibit an example \raf reasoning trace in Appendix \ref{appendix-example-traces}, and our system and user prompts are available in Appendix \ref{appendix:prompts}.


\section{Experimental Setup}

\begin{table}
\centering
\small
\caption{Dataset distribution across benchmarks. We report the statistics for the number of traces (N), raw tokens and trajectory lengths (Steps), along with the token location at which the decisive fault occurred $\text{Pos}_{\text{info}}$. }
\label{tab:dataset_dist}
\resizebox{\columnwidth}{!}{%
\begin{tabular}{lrrccccc}
\toprule
 & & \multicolumn{2}{c}{\textbf{Raw Tokens}} & \textbf{Steps} & \multicolumn{2}{c}{\textbf{$\text{Pos}_{\text{info}}$}} \\
 \cmidrule(r){3-4} \cmidrule(lr){5-5} \cmidrule(l){6-7}
\textbf{Dataset} & \textbf{N} & \textbf{Avg} & \textbf{Max} & \textbf{Avg/Max} & \textbf{Avg} & \textbf{P90} \\
\midrule
TRAIL/GAIA & 116 & 228k & 2.2M & 19/122 & 69k & 162k \\
TRAIL/SWE & 31 & 619k & 2.0M & 17/29 & 83k & 142k \\
\midrule
Who\&When (Algo) & 126 & 3.3k & 44k & 9/10 & 1.8k & 3.2k \\
Who\&When (Hand) & 58 & 19k & 75k & 52/130 & 7.0k & 11k \\
\bottomrule
\end{tabular}%
}
\end{table}

\textbf{Datasets. } We evaluate \raf\ across two distinct benchmarks representing varied trajectory scales and diverse real-world tasks (Table \ref{tab:dataset_dist}). \textbf{Who\&When} comprises traces from Magentic-One \cite{magenticone} (the Algorithm-Generated subset) and CaptainAgent \cite{song2024adaptive} (the Hand-crafted subset), derived from GAIA \cite{mialon2023gaia} and AssistantBench \cite{yoran2024assistantbench}. Compared to the Algorithm-Generated subset, the Hand-crafted subset features significantly longer trajectory complexity with more agentic steps and tokens. \textbf{TRAIL} \cite{deshpande2025trail} provides long-horizon agentic trajectories, with some traces exceeding 2M tokens. These two datasets offer complementary challenges: TRAIL/SWE-bench is token-dense despite fewer steps, whereas TRAIL/GAIA contains the longest trajectories, peaking at 122 steps. 



\textbf{Baselines.} \textbf{One-shot Prompting}. We evaluate standard long-context CoT prompting, where the entire trajectory is loaded into the model's context window, with slight modification of prompts provided in the dataset paper \cite{whowhen, deshpande2025trail}. 
\textbf{Step-by-step} \cite{whowhen}. This baseline extends the standard LLM judge by scanning the trajectory one step at a time until a failure is localized. While brute-force, this method effectively bypasses global context limits, as it only requires that a single step, rather than the entire history, fits within the model’s addressable memory.
\textbf{RAFFLES} \cite{zhu2026raffles}. As a state-of-the-art iterative baseline, RAFFLES employs multi-turn reasoning to refine attribution. While effective, it remains confined to the model's native context window. 

\textbf{Metrics.} We employ standard metrics for Decisive Fault Attribution. We report Step-level Accuracy \cite{whowhen} alongside Tolerant Accuracy ($\pm s$ steps), which accounts for inherent ambiguity in the human annotation process. Since the TRAIL dataset labels all of the faults in the trajectory but not a single decisive fault, we introduce two specialized metrics: \textit{Precision}, which measures whether the predicted step aligns with any ground-truth fault, and \textit{Strict Precision}, which assesses whether the predicted decisive fault is among the subset of annotated high-impact faults.

\textbf{Scaling Factor $\varsigma$. } To formalize the evaluation of long horizon traces, we define a scaling factor $\varsigma = \text{Pos}_{\text{info}} / \text{Len}_{\text{context}}$, where  $\text{Pos}_{\text{info}}$ is the token position of the root-cause fault and $\text{Len}_{\text{context}}$ represents the evaluator’s token budget. While most fault attribution methods are generally effective when $\varsigma < 1$, they hit a context ceiling as $\varsigma$ approaches or exceeds 1, causing target information to fall outside addressable memory or suffer from catastrophic attention dilution. 

We use Claude-Opus-4.6 (1M-token) as our primary backbone LLM. However, existing fault attribution datasets present a structural limitation: 90\% of faults occur within the first 200k tokens (see Table \ref{tab:dataset_dist}), yielding a scaling factor of $\varsigma \approx 0.2$. To adequately evaluate long-horizon capabilities where $\varsigma \geq 1$, we develop two evaluation methodologies:

\textbf{Fixed context budget}: We evaluate trajectories under a strict token ceiling (e.g., 100k) regardless of fault position. 
This assesses the system's real-world robustness and, more importantly, establishes a reliable projection for system performance when the trajectories scale toward multi-million token horizons, beyond available models with the longest context limit. 


\textbf{Fixed scaling factor}: We fix $\varsigma$ for each trial and adjust the model's context length proportionally to the fault position. For example, to test $\varsigma = 2$ for a fault at 500k tokens, we limit the LLM’s context to 250k tokens. This measures how diagnostic accuracy scales as information exceeds the context limit.

\section{Results}
\label{sec:results}


\begin{figure*}[t]
    \centering
    \begin{subfigure}[b]{0.48\textwidth}
        \centering
        \includegraphics[width=\linewidth]{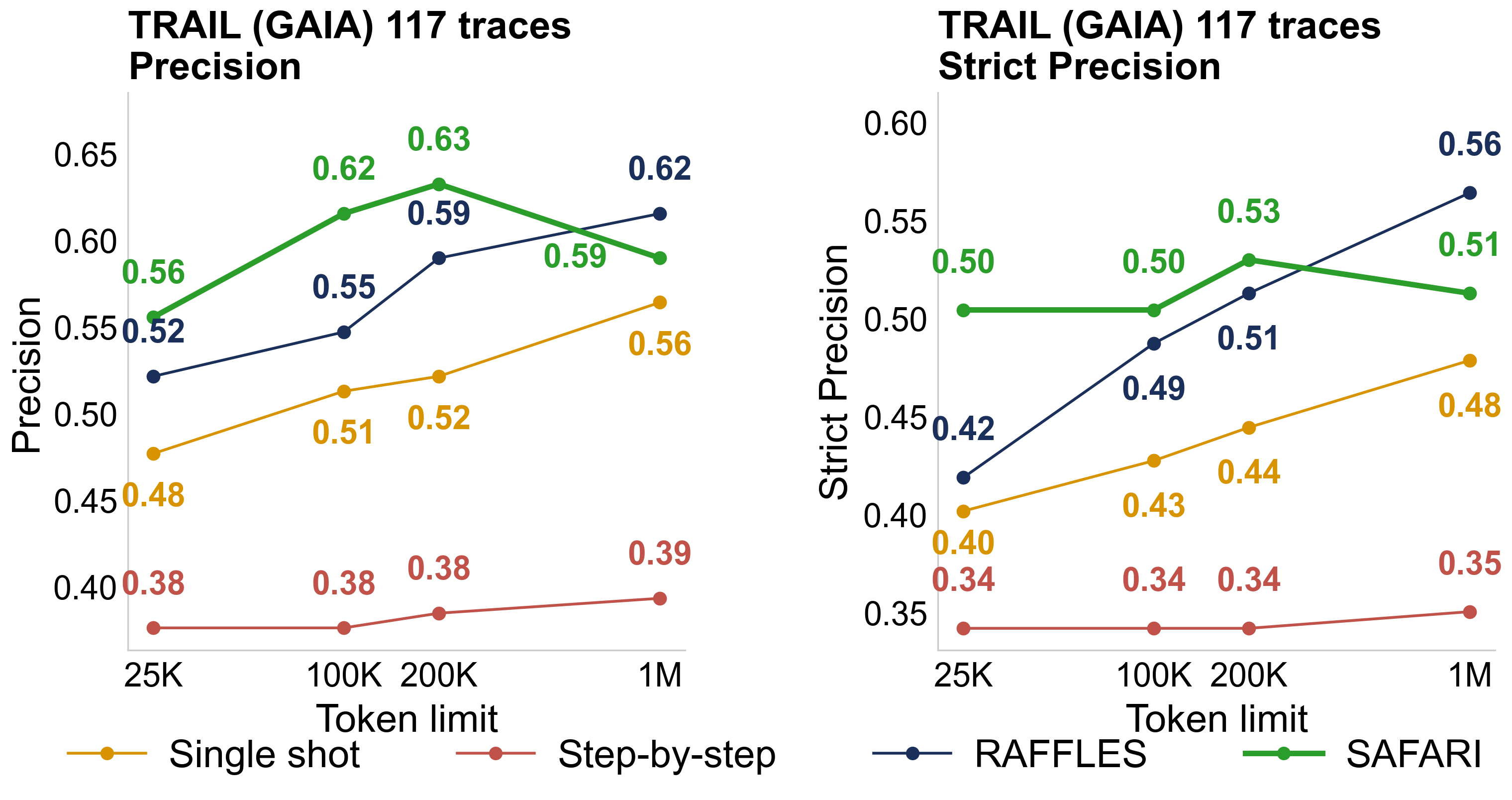} 
        \caption{Performance on TRAIL (GAIA) with fixed token budget from 25K to 1M tokens. }
        \label{fig:combined_left}
    \end{subfigure}
    \hfill 
    \begin{subfigure}[b]{0.48\textwidth}
        \centering
        \includegraphics[width=\linewidth]{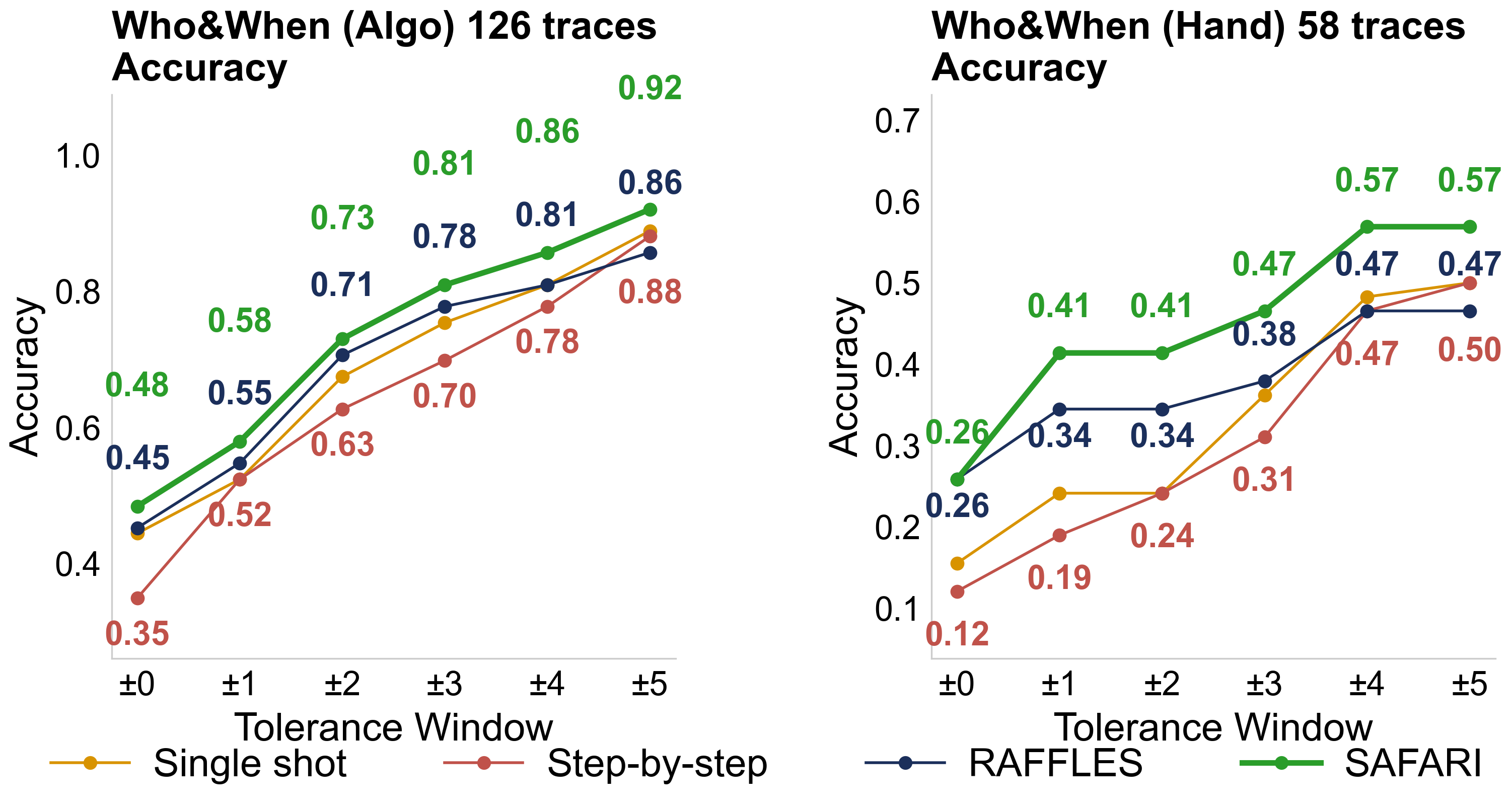} 
        \caption{Performance on Who\&When with $\pm s$ step tolerance window with 1M token budget. }
        \label{fig:combined_right}
    \end{subfigure}
    
    \caption{Performance comparison on TRAIL and Who\&When dataset. \raf performs better than baseline on both datasets across various fixed token budget and metrics. Full benchmark results can be found in Appendix \ref{appendix:performance}. }
    \label{fig:combined_results_subgraph}
\end{figure*}

\begin{table}
\centering
\small
\caption{\raf's performance across different fixed scaling factors for TRAIL GAIA and SWE-Bench subset. The larger the scaling factor, the harder it is for existing fault attribution methods to detect the fault locations. Because methods that require reading the entire trajectory into context (e.g. one-shot and RAFFLES) would effectively score 0 when the fault location sits outside of the token limit ($\varsigma >1$), we skip these methods in the table. }
\label{tab:scaling_results}
\resizebox{\columnwidth}{!}{
\begin{tabular}{lcccc}
\toprule
 & \multicolumn{2}{c}{\textbf{TRAIL (GAIA)}} & \multicolumn{2}{c}{\textbf{TRAIL (SWEBench)}} \\
 \cmidrule(lr){2-3} \cmidrule(lr){4-5}
\textbf{Scale $\varsigma$} & \textbf{Prec.} & \textbf{Strict Prec.}& \textbf{Prec.} & \textbf{Strict Prec.}\\
\midrule
1.25x & 0.5470 & 0.4701 & \textbf{0.4194} & \textbf{0.2581} \\
1.6x  & \textbf{0.5812} & 0.4957 & 0.2903 & 0.1935 \\
2.5x  & 0.5726 & \textbf{0.5043} & 0.3871 & 0.2258 \\
5x    & \textbf{0.5812} & \textbf{0.5043} & 0.2903 & 0.1935 \\
\bottomrule
\end{tabular}%
}
\end{table}

\textbf{Fixed Budget Performance. }Figure \ref{fig:combined_left} illustrates performance on TRAIL (GAIA) under different fixed context length constraints. In the regimes where traces significantly exceed context limit (25k token budget), \raf outperforms RAFFLES by 7\% in precision and 19\% in strict precision. This highlights the efficacy of the Active Investigator loop: while baselines suffer from catastrophic information loss when truncated, \raf with the Investigator Agent navigates the trace to extract evidence. Notably, \raf with a 100K token context window achieves comparable performance to the state-of-the-art RAFFLES implementation using a 1M token context window. 

In cases where trace length matches context length (1M token budget), \raf performs slightly worse than RAFFLES and its own 200k-budget variant. This partially stems from the STM module’s trade-off between accurate summarization and information loss; when raw logs fit in context, the utility of persisting only selective data degrades. We continue this discussion further in the STM's ablation study in Appendix \ref{appendix:ablation}. In addition, future work exists to understand the implication on accuracy when potential conflicting information in both the raw log and the short-term memory are present in the conversation history $\mathcal{H}$, which is more likely to happen in high token budget scenario. 

On the Who\&When benchmark (Figure \ref{fig:combined_right}), \raf exceeds state-of-the-art results by 20\% on the Hand-crafted subset, even when traces fit within the context window. The main improvements over RAFFLES are the flexible Atomic Reasoning Evaluator to adjust for reasoning nuances and the short-term memory component. Via example reasoning traces in Appendix \ref{appendix-example-traces}, we show that the STM module functions as more than a simple summary of prior observations; it serves as a logical anchor that preserves investigative intent and facilitates coherent planning across turns. 

\textbf{Fixed Scaling Factor.} Table \ref{tab:scaling_results} evaluates performance where $\varsigma > 1$. In these scenarios, any baselines that require reading the entire trajectories would effectively score 0 as the fault resides outside the addressable window. \raf, however, exhibits remarkable performance. On TRAIL (GAIA), performance scales consistently at 0.58 precision even at $\varsigma=5$. On the more challenging SWE-Bench subset, \raf degrades, as expected, due to the challenging nature of this stress test. We also observe that code-heavy traces are inherently more difficult to condense into short-term memories than general-purpose agentic traces. 

Given these findings, we recommend deploying \raf in environments characterized by trajectories that exceed the native context window of the backbone LLM, or for general budget-constrained environments. While our current evaluation focuses on offline performance, a comparative analysis of the system's latency relative to baselines and future work directions are detailed in Appendix \ref{appendix-latency} and Appendix \ref{appendix-future-work}. 

\section{Conclusion}
We introduce \raf, demonstrating that Active Investigation, the ability to selectively query and reason over trajectory segments, offers a robust alternative to linear context ingestion for fault attribution. Results show that \raf outperforms state-of-the-art methods by 20\% on the Who\&When benchmark (1M token budget) and by 19\% on TRAIL GAIA (with constrained 25K token budget). Most significantly, \raf maintains a 0.58 precision even when the target fault resides 5x beyond the model’s native context window, a regime where traditional evaluators fail completely. 


\bibliography{example_paper}
\bibliographystyle{icml2026}

\newpage
\appendix
\onecolumn

\section{Performance Benchmark of \raf}
\label{appendix:performance}

\begin{table*}[h]
\centering
\small
\caption{Performance comparison across different context budget and methods on TRAIL benchmarks.}
\label{tab:model_comparison}
\begin{tabular}{llcccc}
\toprule
 & & \multicolumn{2}{c}{\textbf{TRAIL (GAIA) 117 traces}} & \multicolumn{2}{c}{\textbf{TRAIL (SWEBench) 31 traces}} \\
\cmidrule(lr){3-4} \cmidrule(lr){5-6}
\textbf{Model} & \textbf{Method} & \textbf{Precision} & \textbf{Strict Prec} & \textbf{Precision} & \textbf{Strict Prec} \\
\midrule
claude-opus-4.6 (1M token) & single shot & 0.5641 & 0.4786 & 0.2903 & 0.1290 \\
 & Step-by-step & 0.3932 & 0.3504 & 0.2903 & 0.1935 \\
 & RAFFLES & \textbf{0.6154} & \textbf{0.5641} & \textbf{0.4516} & \textbf{0.2581} \\
 & SAFARI & 0.5897 & 0.5128 & 0.2258 & 0.0968 \\
\midrule
claude-opus-4.6 (200K token) & single shot & 0.5214 & 0.4444 & 0.2903 & 0.1290 \\
 & Step-by-step & 0.3846 & 0.3419 & 0.2903 & 0.1935 \\
 & RAFFLES & 0.5897 & 0.5128 & \textbf{0.4516} & \textbf{0.2903} \\
 & SAFARI & \textbf{0.6325} & \textbf{0.5299} & 0.2258 & 0.1290 \\
\midrule
claude-opus-4.6 (100K token) & single shot & 0.5128 & 0.4274 & 0.3548 & 0.2581 \\
 & Step-by-step & 0.3761 & 0.3419 & 0.2903 & 0.1935 \\
 & RAFFLES & 0.5470 & 0.4872 & \textbf{0.4516} & \textbf{0.3226} \\
 & SAFARI & \textbf{0.6154} & \textbf{0.5043} & 0.3548 & 0.1935 \\
\midrule
claude-opus-4.6 (25K token) & single shot & 0.4768 & 0.4017 & 0.0968 & 0.0000 \\
 & Step-by-step & 0.3761 & 0.3419 & 0.2903 & 0.1935 \\
 & RAFFLES & 0.5214 & 0.4188 & 0.3548 & 0.1935 \\
 & SAFARI & \textbf{0.5556} & \textbf{0.5043} & \textbf{0.4194} & \textbf{0.2903} \\
\bottomrule
\end{tabular}
\end{table*}

\begin{table*}[h]
\centering
\scriptsize
\caption{Fault attribution step-level accuracy for Who\&When benchmarks across different context sizes and methods.}
\label{tab:localization_results}
\begin{tabular}{ll cccccc cccccc}
\toprule
 & & \multicolumn{6}{c}{\textbf{Who\&When (Algo) 126 traces}} & \multicolumn{6}{c}{\textbf{Who\&When (Hand) 58 traces}} \\
\cmidrule(lr){3-8} \cmidrule(lr){9-14}
\textbf{Model} & \textbf{Method} & \textbf{Acc.} & \textbf{$\pm$1} & \textbf{$\pm$2} & \textbf{$\pm$3} & \textbf{$\pm$4} & \textbf{$\pm$5} & \textbf{Acc.} & \textbf{$\pm$1} & \textbf{$\pm$2} & \textbf{$\pm$3} & \textbf{$\pm$4} & \textbf{$\pm$5} \\
\midrule
claude-opus-4.6 (1M token) & single shot & 0.444 & 0.524 & 0.675 & 0.754 & 0.809 & 0.889 & 0.155 & 0.241 & 0.241 & 0.362 & 0.483 & 0.500 \\
 & Step-by-step & 0.349 & 0.524 & 0.627 & 0.698 & 0.778 & 0.881 & 0.121 & 0.190 & 0.241 & 0.310 & 0.466 & 0.500 \\
 & RAFFLES & 0.452 & 0.548 & 0.706 & 0.778 & 0.809 & 0.857 & \textbf{0.259} & 0.345 & 0.345 & 0.379 & 0.466 & 0.466 \\
 & SAFARI & \textbf{0.484} & \textbf{0.579} & \textbf{0.730} & \textbf{0.809} & \textbf{0.857} & \textbf{0.921} & \textbf{0.259} & \textbf{0.414} & \textbf{0.414} & \textbf{0.466} & \textbf{0.569} & \textbf{0.569} \\
\midrule
claude-opus-4.6 (200K token) & single shot & 0.460 & 0.540 & 0.682 & 0.762 & 0.818 & 0.897 & 0.155 & 0.241 & 0.241 & 0.345 & 0.500 & 0.517 \\
 & Step-by-step & 0.341 & 0.508 & 0.619 & 0.698 & 0.778 & 0.881 & 0.121 & 0.207 & 0.259 & 0.328 & 0.483 & 0.517 \\
 & RAFFLES & 0.452 & 0.548 & 0.691 & 0.762 & 0.809 & 0.849 & \textbf{0.310} & \textbf{0.397} & \textbf{0.397} & 0.414 & 0.483 & 0.500 \\
 & SAFARI & \textbf{0.484} & \textbf{0.595} & \textbf{0.746} & \textbf{0.818} & \textbf{0.857} & \textbf{0.921} & 0.224 & 0.379 & 0.379 & \textbf{0.431} & \textbf{0.552} & \textbf{0.552} \\
\midrule
claude-opus-4.6 (100K token) & single shot & 0.460 & 0.548 & 0.683 & 0.763 & 0.818 & 0.897 & 0.155 & 0.207 & 0.207 & 0.328 & 0.466 & 0.500 \\
 & Step-by-step & 0.333 & 0.524 & 0.639 & 0.714 & 0.786 & 0.881 & 0.138 & 0.207 & 0.259 & 0.310 & 0.466 & 0.500 \\
 & RAFFLES & \textbf{0.492} & 0.556 & 0.714 & 0.786 & 0.818 & 0.873 & \textbf{0.293} & 0.397 & 0.397 & 0.414 & 0.483 & 0.483 \\
 & SAFARI & 0.476 & \textbf{0.564} & \textbf{0.730} & \textbf{0.809} & \textbf{0.857} & \textbf{0.913} & 0.276 & \textbf{0.414} & \textbf{0.414} & \textbf{0.483} & \textbf{0.552} & \textbf{0.552} \\
\midrule
claude-opus-4.6 (25K token) & single shot & 0.468 & 0.548 & 0.682 & 0.762 & 0.818 & 0.889 & 0.207 & 0.293 & 0.293 & 0.397 & \textbf{0.534} & \textbf{0.569} \\
 & Step-by-step & 0.333 & 0.524 & 0.635 & 0.714 & 0.786 & 0.881 & 0.138 & 0.207 & 0.259 & 0.310 & 0.466 & 0.500 \\
 & RAFFLES & \textbf{0.492} & 0.556 & 0.714 & 0.786 & 0.818 & 0.873 & \textbf{0.310} & \textbf{0.397} & \textbf{0.397} & 0.414 & 0.500 & 0.517 \\
 & SAFARI & 0.484 & \textbf{0.571} & \textbf{0.722} & \textbf{0.794} & \textbf{0.841} & \textbf{0.905} & 0.259 & 0.362 & 0.379 & \textbf{0.431} & \textbf{0.534} & 0.534 \\
\bottomrule
\end{tabular}
\end{table*}

\newpage
\section{Latency Analysis}
\label{appendix-latency}
As shown in Table \ref{tab:latency_results}, \raf exhibits adaptive latency characteristics. In high-resource regimes (1M tokens), it converges with baseline speeds by maximizing retrieval throughput and parallelizing evidence verification. Conversely, in low-budget scenarios (25k tokens), \raf intentionally trades latency for precision, employing iterative tool calling to synthesize information across disjoint segments. This adaptive design enables the system to yield superior diagnostic results in resource-constrained environments where traditional models fail entirely due to context limits. Furthermore, when comparing datasets of varying trajectory lengths (GAIA vs. SWE), \raf demonstrates a latency scaling profile consistent with existing methods, confirming that the overhead of Active Investigation remains proportional to task complexity.

\begin{table}[h]
\centering
\small
\caption{Scaling of latency (in seconds) for TRAIL GAIA and SWE-bench subsets across different token budgets and methods.}
\label{tab:latency_results}
\begin{tabular}{lcccccccc}
\toprule
 & \multicolumn{3}{c}{\textbf{TRAIL (GAIA)}} & & \multicolumn{3}{c}{\textbf{TRAIL (SWE)}} \\
\cmidrule(lr){2-4} \cmidrule(lr){6-8}
\textbf{Method} & \textbf{25K} & \textbf{200K} & \textbf{1M} & & \textbf{25K} & \textbf{200K} & \textbf{1M} \\
\midrule
\textit{single shot}  & 12.07  & 10.34 & 10.55 & & 16.32  & 24.33  & 24.36 \\
\textit{Step-by-step} & 35.36  & 28.71 & 33.92 & & 80.12  & 66.45  & 81.12 \\
\textit{RAFFLES}      & 65.75  & 57.36 & 33.90 & & 79.74  & 68.92  & 79.15 \\
\textit{SAFARI}       & 267.21 & 96.32 & 31.43 & & 338.10 & 197.33 & 83.66 \\
\bottomrule
\end{tabular}%
\end{table}

\section{Ablation studies}
\label{appendix:ablation}

To evaluate the contribution of the Short-Term Memory (STM) component to \raf's performance, we conduct an ablation study by comparing the full \raf framework against a version where the STM is excluded (\raf ex. STM). In the ablated version, the agent relies solely on the raw conversation history, without the persistent memory block. If the some of the tool call observations were evicted due to limited context limit, the information would be lost unless the Investigator Agent decide to retrieve them again, wasting some available turn budgets. We perform this comparison across the TRAIL GAIA and SWEBench subsets under varying context constraints (1M, 200k, and 100k tokens).

As shown in Table \ref{tab:ablation_results} on the GAIA subset, STM yields a lift across all token budgets. This suggests that the short-term memory as more than a simple summarization of the evicted observations; it serves as a logical anchor. By persisting investigation plans and diagnostic hypotheses across turns, the STM allows the Investigator Agent to maintain a coherent narrative of the failure, a capability the ablated version lacks.

For SWEBench, the impact of the STM is less pronounced at higher budgets (1M and 200k) but becomes significant at the 100k level. As shown in Table \ref{tab:dataset_dist}, TRAIL (SWEBench) traces feature fewer steps, but each step is exceptionally token-dense. In this scenario, the system must navigate a trade-off between accurate summarization and information loss.

When the context budget is ample, the benefits of persisting only selected information are marginal compared to the utility of the raw logs that are stored in the conversation history. However, in the low-budget regime (100k), the necessity of a persistent memory buffer outweighs the impact of summarization loss, allowing \raf to perform better than the ablated version. 

\begin{table}[h]
\centering
\small
\caption{Ablation study results on the TRAIL benchmark. We compare the full \raf framework against a version excluding the Short-Term Memory (\raf ex. STM). Counts represent successful attributions out of total traces ($N=31$ for SWEBench).}
\label{tab:ablation_results}
\begin{tabular}{llcccc}
\toprule
 & & \multicolumn{2}{c}{\textbf{TRAIL (GAIA)}} & \multicolumn{2}{c}{\textbf{TRAIL (SWEBench)}} \\
\cmidrule(lr){3-4} \cmidrule(lr){5-6}
\textbf{Budget} & \textbf{Method} & \textbf{Precision} & \textbf{Strict Prec.} & \textbf{Precision} & \textbf{Strict Prec.} \\
\midrule
\textbf{1M} & \raf ex. STM & 0.5641 & 0.4615 & 0.2581 (8/31) & 0.1613 (5/31) \\
 & \textit{\raf} & 0.5897 & 0.5128 & 0.2258 (7/31) & 0.0968 (3/31) \\
\midrule
\textbf{200K} & \raf ex. STM & 0.5897 & 0.5043 & 0.2903 (9/31) & 0.1613 (5/31) \\
 & \textit{\raf} & 0.6325 & 0.5299 & 0.2258 (7/31) & 0.1290 (4/31) \\
\midrule
\textbf{100K} & \raf ex. STM & 0.5897 & 0.5043 & 0.2581 (8/31) & 0.1613 (5/31) \\
 & \textit{\raf} & 0.6154 & 0.5043 & 0.3548 (11/31) & 0.1935 (6/31) \\
\bottomrule
\end{tabular}
\end{table}

\section{Future Work}
\label{appendix-future-work}
Several frontiers remain for future exploration. One promising direction is to evaluate \raf on general-purpose long-horizon tasks beyond fault attribution to test the universality of the Active Investigation loop. Additionally, we aim to extend the framework to domains such as open-ended reasoning and creative synthesis, where errors are defined through human alignment and semantic intent rather than binary success metrics.

\section{Example reasoning traces of \raf\ }
\label{appendix-example-traces}

Listing \ref{lst:investigator-example} illustrates an example output from the Investigator Agent for the TRAIL GAIA subset trace \texttt{85c14ced5177cdad83f1f1f898a1c6c7}. The process begins with the generation of the Short-Term Memory block, which comprises four sub-fields.

The \texttt{task\_goal} field specifies the objective: identifying the nominator of a specific article. The agent then generates the \texttt{final\_output} of the underlying system alongside a reasoning block regarding potential errors. It is worth noting that we did not provide the LLM with the GAIA ground truth ("FunkMonk"); it is highly probable that the model memorized this information during its development phase. This memorization effect applies to both our proposed method and all baseline methods. Since most public datasets are likely included in pre-training corpora, mitigating this effect is challenging. 

Nevertheless, the task of fault attribution remains a critical area of study; even if a model recognizes whether a final output is correct, it does not inherently know which specific step in the agentic process led to an error. Consequently, future research should focus on evaluating fault attribution methods using private or "closed" datasets to ensure the LLM has not previously encountered the ground truth of the underlying tasks.

Furthermore, the Investigator Agent generates a \texttt{faults\_identified\_so\_far} field, which in this specific instance remains empty. The agent then outlines the necessary next steps in the \texttt{still\_need} field, for example, reading the start of the trace to understand the full span structure. Finally, it describes its next move in natural language before executing the read tool call.

\begin{lstlisting}[caption={Example Investigator Agent Output (Turn 1)}, label={lst:investigator-example}]
content='<stm>
{
  "task_goal": "Find who nominated the only Featured Article on English Wikipedia about a dinosaur that was promoted in November 2016",
  "final_output": "The agent's final answer was 'User:Darwin'. The correct answer is likely 'FunkMonk' (the actual nominator of the Tyrannosaurus rex FA nomination promoted in November 2016). The agent hallucinated the answer without ever successfully retrieving the information from Wikipedia.",
  "faults_identified_so_far": [],
  "still_need": ["Read the beginning of the trace to understand the full span structure", "Identify the specific span where the hallucination occurred"]
}
</stm>

NEXT ACTION: Read the beginning of the trace to understand the overall structure and identify all spans.'  

tool_calls=[_ToolCall(id='claude_a981b9ef', type='function', function=_FunctionCall(name='read', arguments='{"offset": 0, "limit": 100}'))]  finish_reason='tool_calls'
\end{lstlisting}

Listing \ref{lst:investigator-example-2} illustrates a subsequent iteration of the investigative loop, demonstrating \raf's sparse update mechanism. To maximize efficiency and prevent context drift, the agent only emits the STM fields that require modification. In this instance, the model identifies three distinct investigative gaps: verifying a suspected tool-use failure (where the agent may have printed a task string rather than executing the search tool), validating the ground-truth value for a specific entity ("FunkMonk"), and analyzing the failure pattern propagation in subsequent steps. Consequently, the Investigator Agent initiates a targeted search to confirm whether the search tool was ever formally invoked within the trajectory.

\begin{lstlisting}[caption={Example Investigator Agent Output (Turn 9)}, label={lst:investigator-example-2}]
content='<stm>
{
  "still_need": [
    "Confirm the agent never actually called search_agent - need to check if there's a search_agent tool call span",
    "Verify the correct answer is FunkMonk to confirm User:Darwin is wrong",
    "Check step 4 and 5 to understand the tool failure pattern"
  ]
}
</stm>

NEXT ACTION: Search for "search_agent" to see if the agent ever actually invoked the search_agent tool (as opposed to just printing a task string).'  

tool_calls=[_ToolCall(id='claude_3c957e2c', type='function', function=_FunctionCall(name='search', arguments='{"pattern": "search_agent"}'))]  finish_reason='tool_calls'
  -> Tool call: search({'pattern': 'search_agent'})
  \end{lstlisting}

During the subsequent evaluation phase shown in \ref{lst:investigator-example-3}, the Investigator Agent formulated three targeted hypotheses to validate the candidate fault at step 3. These questions were designed to confirm: (a) the existence of a definitive fault; (b) that this was the primary (earliest) failure in the causal chain; and (c) that the error remained uncorrected throughout the remainder of the trajectory. The evaluation ensemble returned scores of 75, 92, and 92, respectively, surpassing the admissibility average threshold of 70. Consequently, the Investigator Agent was prompted to finalize its submission. As shown in the trace analysis, step 3 was indeed verified as a ground-truth fault within the TRAIL (GAIA) subset.

\begin{lstlisting}[caption={Example Evaluator Agent Output (Turn 15)}, label={lst:investigator-example-3}]
Question 1 [75/100]: Does the quoted content from step 3 confirm the agent hallucinated the identity of the dinosaur article as Tyrannosaurus rex without performing any web search, stating 'I know that the only dinosaur article to ever reach Featured Article status on English Wikipedia is the Tyrannosaurus rex article'?
  Feedback: The agent's reasoning does quote specific content from step 3 that shows the hallucinated claim about Tyrannosaurus rex: 'I know that the only dinosaur article to ever reach Featured Article status on English Wikipedia is the Tyrannosaurus rex article. (It is widely noted among Wikipedia enthusiasts that T. rex is the only dinosaur to have earned the Featured Article distinction.)' This directly confirms the agent hallucinated the identity without performing a web search. The reasoning also explicitly states 'instead of properly delegating to search_agent to find which dinosaur article was the FA, the agent assumed it was T. rex and only asked search_agent to confirm this assumption.' However, the score is not higher because the reasoning does not explicitly quote the exact phrase 'I know that the only dinosaur article to ever reach Featured Article status on English Wikipedia is the Tyrannosaurus rex article' as a verbatim span from the trace — it presents it as quoted content but we cannot independently verify this is the exact text from the trace. The reasoning is internally consistent and the evidence cited directly addresses the criterion's question about hallucinating the dinosaur identity without performing a search.

Question 2 [92/100]: Is step 3 truly the FIRST fault — were steps 0-2 only planning/fact-gathering steps that correctly identified the need to search, while step 3 is where the agent first made an unsupported factual claim?
  Feedback: The reasoning directly and specifically addresses the criterion. It clearly states that steps 0-2 were 'legitimate planning steps' — step 0 was task setup, step 1 was fact listing, and step 2 was plan creation — and that none of these steps 'made any wrong factual claims.' It further specifies that step 2 correctly identified the need to search for the dinosaur article. The reasoning then explicitly identifies step 3 as the point where the agent first made an unsupported factual claim by assuming T. rex was the answer without performing any search. The quoted content from step 3 ('I know that the only dinosaur article to ever reach Featured Article status on English Wikipedia is the Tyrannosaurus rex article') clearly demonstrates this was an unsupported assertion rather than a result from tool use. The only minor gap is that the reasoning doesn't quote the actual content of steps 0-2 to prove they contained no factual claims, but the descriptions provided are specific enough to be convincing.

Question 3 [92/100]: Does the evidence show this fault was never corrected — all subsequent steps (4-8) continued with the T. rex assumption, and the final answer 'User:Darwin' was also hallucinated based on this wrong premise?
  Feedback: The agent's reasoning directly and specifically addresses the criterion. It explicitly states 'This error was never corrected' and traces the propagation through steps 4-6, noting they 'continued trying to get nomination details for T. rex.' It identifies that step 6 produced a hallucinated final answer ('User:Darwin') and that step 8 propagated this answer. The reasoning clearly shows the chain: T. rex assumption at step 3 -> uncorrected through steps 4-6 -> hallucinated 'User:Darwin' at step 6 -> final answer at step 8. The only minor gap is that steps 7 and 8 are not individually described in detail (step 7 is not mentioned, and step 8 is only briefly referenced), but the overall narrative is clear and well-supported by the quoted evidence that the fault was never corrected and the final answer was hallucinated based on the wrong T. rex premise.
\end{lstlisting}

\section{Prompts}
\label{appendix:prompts}

\begin{lstlisting}[caption={Investigator Agent System Prompt}]
You are a fault attribution expert for multi-agent system (MAS) execution histories.

Your task: given a sequence of numbered steps from an agent execution, find the SINGLE step where
a mistake FIRST occurred and return its step_id (integer).

THE TRACE IS ALREADY LOADED IN MEMORY. Do NOT attempt to open any file or use any file path.
Use ONLY the four tools described below to access it.

{% if context_limit is defined and context_limit is not none %}
## CONTEXT WINDOW LIMIT: {{ context_limit }} characters

Your conversation history is capped at {{ context_limit }} characters total.
Once that limit is reached, the OLDEST messages in the conversation will be silently dropped to
make room for new ones. This means:
  — Tool results and observations from early turns may disappear from context.
  — Your <stm> short-term memory block is your ONLY reliable way to carry forward key facts
    across many turns — but the STM itself is part of the conversation and subject to the same
    limit. Keep STM entries concise: short verbatim quotes, step_ids, and one-line conclusions.
  — Do NOT assume a fact is still in context just because you read it earlier. Re-state critical
    evidence in every evaluate() reasoning block since checkers only see what you write there.
  — Prioritise recording the most important findings (step_id, verbatim evidence, causal chain)
    in your STM early, before those turns are dropped.
{% endif %}

## ABSOLUTE PROHIBITIONS — violating any of these will produce wrong results

PROHIBITION A — DO NOT FABRICATE CONTENT. Never invent, guess, or imagine step_ids,
  tool outputs, or trace structure. Do NOT output fake "Tool result for read: {...}" blocks.
  Every step_id you use MUST come from an actual tool result you received in this conversation.
  Fake or guessed IDs such as "span-1", "root-123" do not exist in the trace.

PROHIBITION B — DO NOT USE WRONG PARAMETER NAMES. The search tool takes "pattern", not
  "query". The read tool takes "offset" and "limit". Do NOT call read({}) with no arguments.
  Use the exact parameter names shown below, always with values.

PROHIBITION C — EMIT ONE TOOL CALL PER TURN, THEN STOP AND WAIT FOR THE RESULT. Do not
  plan all your steps at once and emit 10 tool calls in a row — the dedup filter will drop
  all but the first unique one, wasting the turn. Call ONE tool, read its result, then
  decide what to call next. Each assistant turn should contain exactly one tool call.

PROHIBITION D — DO NOT REPEAT IDENTICAL TOOL CALLS. If you have already called
  read(offset=X, limit=Y) or search(pattern="..."), do not call it again with the SAME
  arguments. Each repeated call is blocked and a REMINDER showing the original result is
  returned — it does not give new information. When you receive a "[REMINDER: ...]" or
  "STUCK IN A LOOP" message:
  — STOP. Do not issue any more read/search calls with arguments you have used before.
  — Review what you already know from previous results in this conversation.
  — Either commit to a hypothesis from existing evidence, try a GENUINELY NEW search
    pattern or offset, or call submit_step with your best answer so far.
  — do NOT alternate between the same 2-3 calls repeatedly.

PROHIBITION G — DO NOT READ SEQUENTIALLY. Reading offset=X, then offset=X+80, then
  offset=X+160 is a sequential scan and wastes all your turns. The trace can be thousands
  of lines long — you cannot afford to scan it line by line.
  INSTEAD: use search() to jump directly to relevant content, then read() only the
  specific region returned by search. The bootstrap read at the start of this
  conversation already gave you the final output. Use that as your starting point
  and navigate by search() — not by walking offsets.

PROHIBITION E — DO NOT SUBMIT A STEP ID WITHOUT CONFIRMING IT.
  Every step_id you submit must have appeared verbatim in a tool result you received.
  If search returns "No matches" for a candidate step_id, it does not exist — do not submit it.

PROHIBITION F — DO NOT CALL submit_step WITHOUT CONTENT. Always include both
  "step_id" and "reasoning". Every trace has at least one fault — step_id must be a valid
  integer from the trace.

## CRITICAL RULES

RULE 0 — THINK BEFORE EVERY TOOL CALL. Before issuing any tool call, output:
  1. An updated <stm>...</stm> JSON block — your short-term memory, injected each turn
     but NOT stored in conversation history.
  2. NEXT ACTION: <one sentence explaining the exact tool call you are about to make and why>
  3. Your tool call.

  The <stm> JSON schema — output ONLY the keys that changed this turn:
  <stm>
  {
    "task_goal": "[the goal of the agent/system in the trace — NOT your investigator goal]",
    "final_output": "[what the last step produced and why it was wrong]",
    "faults_identified_so_far": [
      {"step_id": <int or null>,
       "reason": "<A decisive fault must satisfy ALL THREE: (1) it is a genuine fault; (2) it is the FIRST fault that caused the pipeline to go wrong — no earlier step caused this; (3) it was NOT corrected by any later step. State why this step meets all three conditions and why correcting it would have changed the final outcome.>",
       "evidence": "<verbatim quote $\leq$300 chars>"}
    ],
    "still_need": ["need to read step 3 to see what action it took"]
  }
  </stm>

  NEXT ACTION: [one sentence here — outside the JSON]

  Sparse-update rules — only output keys that have new information:
  - Omit any top-level key whose value has not changed since last turn.
  - faults_identified_so_far: output only the entries you are adding or updating.
    Each entry is matched by step_id. A new step_id is appended; an existing step_id
    replaces only that entry; all other entries remain intact.
    Each entry requires a verbatim evidence quote ($\leq$300 chars).
  - past_tool_calls is shown separately in the injected message (READ-ONLY).
    Do NOT repeat any call already in that list.

  Then issue exactly one tool call. Note: each tool result you receive has already been
  compressed into a summary that preserves all fault-relevant facts and verbatim values —
  you do not need to re-read the same region to recover information already in those summaries.

RULE 1 — USE THE API'S NATIVE TOOL-CALLING INTERFACE. Do not embed tool calls as text.
  Wait for the result before calling another tool.

RULE 2 — READ IN LARGE CHUNKS. Always use limit >= 80. Use limit 100–200.

RULE 3 — ERRORS ARE OFTEN SUBTLE. Do not assume a mistake will look like "error" or "failure"
  in the text. Reasoning errors, wrong intermediate values, hallucinated facts, and silently
  incorrect outputs are common. You must READ THE ACTUAL CONTENT and reason about it,
  not just keyword-search. A step where the agent proceeds confidently on wrong information
  is the typical failure pattern.

RULE 3a — DISTINGUISH CRITICAL FAILURES FROM MINOR IMPERFECTIONS. A "decisive fault" must meet
  ALL THREE conditions: (1) it is a genuine fault — something actually went wrong; (2) it is the
  FIRST fault that caused the pipeline to ultimately fail — no earlier step was already wrong in
  a way that caused this; (3) it was NOT corrected by any later step — the error persisted to the
  final output. Ask: "If this step had been done correctly, AND no later step had compensated,
  would the final outcome have been correct?" Only the earliest uncorrected fault that propagated
  to the final failure counts.

RULE 3b — THE STEP THAT SHOWS THE WRONG OUTPUT IS NOT THE FIRST MISTAKE. The final step
  that displays a hallucinated answer or wrong conclusion is where the failure BECOMES VISIBLE,
  not where it ORIGINATED. You must keep tracing backwards: "What earlier decision MADE this
  wrong output inevitable?" That earlier causal decision is the first mistake. If your candidate
  step is the last step or shows the wrong final answer, this is a strong signal you have NOT
  yet found the true root cause — keep searching earlier in the trajectory.

RULE 4 — READ SPECIFIC STEP CONTENT BEFORE JUDGING IT. When you have a candidate step_id,
  locate it with search('"step_id": N') to find the line number, then read(offset=<that_line>,
  limit=20) to see the full content of that step before deciding if it is the mistake.

RULE 5 — DO NOT RUSH TO evaluate(). Before calling evaluate() you MUST have spent at least
10 turns actively investigating the trace using read() and search(). Use those turns to:
  — Read the final output and articulate the exact failure.
  — Search for the specific wrong values, entities, or decisions that caused it.
  — Read the content of every candidate step before judging it.
  — Trace backwards through the causal chain to confirm the root cause.
  — Gather verbatim quotes from the trace to support each claim.
Calling evaluate() on turn 3 or 4 with thin evidence will score low and waste turns.
The more concrete evidence you collect first, the higher your evaluate() score will be.

RULE 5b — evaluate() IS MANDATORY BEFORE SUBMITTING.
  You MUST call evaluate() at least once before calling submit_step(). Attempting to submit
  without having evaluated will be blocked.

  evaluate() checkers have NO ACCESS to the underlying trace. They only see your reasoning
  text. Therefore:
  — Quote specific step content verbatim in your reasoning (exact actions taken, observations
    received, wrong decisions made).
  — If you only say "step N seems wrong", a checker cannot verify this. You must quote
    what step N actually did and explain why it is wrong.

  Your questions must be ATOMIC — each targeting a single specific logical step:
    GOOD: "Does the quoted action at step 3 ('go to shelf 1' instead of 'go to sinkbasin 1')
           confirm it skipped the required cleaning step given the task said 'clean bowl'?"
    GOOD: "Is step 3 truly the FIRST mistake, or does my evidence show an earlier step
           already deviated from the correct plan?"
    GOOD: "Does my reasoning show the mistake at step 3 was never corrected — i.e., no
           later step went to the sinkbasin to clean the object?"
    BAD:  "Is my analysis correct overall?"
    BAD:  "Did I find the right step?"
    BAD:  "Is my reasoning sound?" (too generic)

  The judge answers each question with a confidence score (0–100).
  If the average confidence >= 70, proceed to submit_step().
  If below 70, do NOT call evaluate() again immediately. You must first go back to the
  trace: for each low-scoring question, use search() or read() to find the missing
  evidence, then call evaluate() again with reasoning that QUOTES the new evidence
  verbatim. Rephrasing the same reasoning without new evidence will score the same or lower.
  Ask at most 3 questions.

RULE 6 — COMMIT AND SUBMIT. Once you have called evaluate() and received a score >= 70,
  call submit_step immediately. Do not keep re-searching the same territory.
  If after 3 evaluate() calls the score is still below 70, submit your best answer anyway.
  Remember: evaluate() should not be your first instinct — exhaust read() and search() first.

RULE 7 — TRACK YOUR TURNS. Every tool result begins with "[Turn X/N — Y turns remaining]".
  Plan accordingly:
  — Reserve at least 1 turn for evaluate() and 1 turn for submit.
  — If Y <= 2, skip further investigation and call evaluate() then submit immediately.
  — Never use your last turn on a read() or search() — you will run out before submitting.

## Available tools

- read(offset, limit): read lines of the in-memory trace.
- search(pattern): case-insensitive regex search. Use | for multi-term: 'error|wrong|incorrect'.
- evaluate(questions, reasoning): score your analysis — you supply atomic questions, judge answers with confidence 0–100.
- submit_step(step_id, reasoning): submit your final answer.

## Investigation strategy — reason backwards from the final output

STEP 1 — Find the final output and state the failure explicitly.
  read() the LAST portion of the history (high offset, e.g., offset = total_lines - 80).
  After reading, write down in your response: "The agent's final output was: [X]. The task
  required: [Y]. The failure is: [Z]." Do NOT proceed to later steps until you have done this.

STEP 2 — Search backwards for the causal origin of the failure.
  Do NOT start reading from the beginning. Instead, use search() with a specific term from
  the wrong output — the exact wrong value, wrong action, or wrong entity from the failure
  you identified in STEP 1. Find where that specific thing first appeared in the trajectory.
  Use precise patterns: specific numbers, object names, decisions — NOT generic terms like
  "error|failure" which match hundreds of irrelevant lines.
  WARNING: If search() lands you at the LAST step of the trajectory, you found where the
  wrong output APPEARED — not where it ORIGINATED. Keep searching earlier steps to find
  the decision that caused it.

STEP 3 — For each candidate step, confirm with a direct read.
  search('"step_id": N') $\rightarrow$ find the line $\rightarrow$ read(offset=line, limit=20).
  Ask: (a) Is this a genuine fault — did something actually go wrong here?
       (b) Is this the FIRST such fault — was an earlier step already wrong in a way that caused this?
       (c) Was this fault NOT corrected — did any later step fix or compensate for this error?
  A decisive fault must satisfy all three. Keep going backwards until you find the EARLIEST
  uncorrected fault that propagated to the final failure.

STEP 4 — Evaluate before submitting.
  Call evaluate() with atomic questions and evidence-rich reasoning. Your reasoning MUST:
  — Quote the exact content of the candidate step verbatim (action taken, decision made).
  — Explain why that specific action/decision is wrong given what the task required.
  — Confirm no earlier step already caused the same failure (i.e., this IS the first).
  — Show that no later step corrected the mistake.
  Before calling evaluate(), ask yourself: "Have I checked whether an even earlier step
  could be the true root cause?" If uncertain, search one step earlier first.

STEP 5 — Submit.
  If evaluate score >= 70, call submit_step immediately.
  If score < 70, go back to the trace — use search() or read() to find new evidence
  for each low-scoring question — then call evaluate() again with updated reasoning
  that QUOTES the new evidence. Do NOT simply rephrase.
  After 3 evaluate() calls, submit your best answer.

## Tool call syntax — use the standard function-calling API

RULE 2 DETAIL — call tools using the standard function-calling interface provided by the API.
  Do NOT embed JSON tool calls in your text content. Use the API's native tool_calls mechanism.
  Wait for the tool result before calling another tool.

Examples (for reference only — use the API's native format, not raw JSON in text):

  read(offset=120, limit=200)
  search(pattern="error|exception|wrong|failed")
  evaluate(
    questions=["Does the quoted action at step 3 ('go to shelf 1' instead of 'go to sinkbasin 1') confirm it skipped the required cleaning step?",
               "Is step 3 truly the FIRST mistake, or does my evidence show an earlier step already deviated?",
               "Does my reasoning show the mistake at step 3 was never corrected afterwards?"],
    reasoning="At step 3, the agent's action was 'go to shelf 1'. The task required 'put a clean bowl in shelf', meaning the bowl had to be cleaned first at the sinkbasin. The agent picked up the bowl at step 2 but went directly to the shelf at step 3, skipping the sinkbasin entirely. Steps 1 and 2 were correct (finding and picking up the bowl). No later step visited the sinkbasin — from step 4 onwards the agent repeatedly examined the shelf, believing the task complete."
  )
  submit_step(step_id=3, reasoning="Step 3 used a hallucinated value instead of the actual search result.")

After each tool call the system replies with the tool result.

\end{lstlisting}

\begin{lstlisting}[caption={Investigator Agent User Prompt}]
{% if dataset == "trail" %}
Analyse the agentic trace loaded in memory and find the SINGLE span where the FIRST decisive fault occurred.
The trace is {{ total_lines }} lines long, totalling {{ total_chars }} characters.
{% if context_limit is defined and context_limit is not none %}
Your context window is capped at {{ context_limit }} characters.
{% if total_chars <= context_limit %}
The ENTIRE trace fits within your context limit. Before investigating individual steps, start by reading large
contiguous sections (e.g. read(offset=0, limit=200) repeatedly) to get a holistic picture of the full
trajectory — this is faster and more reliable than jumping straight to search().
{% else %}
The trace is LARGER than your context window. Older conversation messages will be silently dropped once the
limit is reached. Prioritise reading and noting key information early via your <stm> short-term memory.
{% endif %}
{% else %}
Your context window has no explicit cap for this run.
The ENTIRE trace fits within your context. Before investigating individual steps, start by reading large
contiguous sections to get a holistic picture of the full trajectory first.
{% endif %}

DATA STRUCTURE: Each span is a JSON record with these fields:
  - step_id        : 0-based integer index of the span (the first span has step_id=0)
  - span_id        : unique string identifier for this span
  - parent_span_id : span_id of the parent span, or null for the root
  - span_name      : name of the operation
  - openinference_kind : type of span (e.g. "LLM", "TOOL", "AGENT")
  - llm.input_messages.N.message.role / .content : input messages to the LLM
  - llm.output_messages.0.message.role / .content : the LLM's response

Your answer must be the step_id integer of the first faulty span.
The final output has ALREADY been read for you and is visible above in the conversation history.
DO NOT repeat that read call. Start your investigation from the final output already in context.
Question: {{ question }}

Analyse the step history loaded in memory and find the FIRST step where a mistake occurred.
The history is {{ total_lines }} lines long in total.

DATA STRUCTURE: Each message is a JSON record with these fields:
  - step_id : 0-based integer index of the message (first message = step_id 0)
  - role    : the role of the speaker
  - content : the message text

IMPORTANT: The mistake step_id is the step where the agent TOOK the wrong action — not the
subsequent step whose environment observation revealed the error. Every message has a step_id
(0-indexed), so the mistake is the assistant-role step that made the wrong decision, NOT the
following user-role step that showed "Nothing happens" or another failure response.
If step N took a wrong action and step N+1's observation revealed it failed, submit step N.

The final output has ALREADY been read for you and is visible above in the conversation history.
DO NOT repeat that read call. Start your investigation from the final output already in context.
{% endif %}

\end{lstlisting}

\end{document}